\documentclass{article}

\usepackage{microtype}
\usepackage{graphicx}
\usepackage{subcaption}
\usepackage{booktabs} 

\usepackage{amsmath}
\usepackage{amssymb}
\usepackage{algorithm}
\usepackage{algpseudocode}

\usepackage{hyperref}

\usepackage[accepted]{icml2026}

\usepackage{mathtools}
\usepackage{amsthm}
\usepackage[table]{xcolor} 
\usepackage{multirow}
\definecolor{pastelblue}{RGB}{179, 205, 227}
\definecolor{pastelpink}{RGB}{245, 205, 205}

\usepackage{listings}
\usepackage[capitalize,noabbrev]{cleveref}

\theoremstyle{plain}

\theoremstyle{definition}

\theoremstyle{remark}

\usepackage[textsize=tiny]{todonotes}

\icmltitlerunning{BiomedXPro: Prompt Optimization for Explainable Diagnosis with Biomedical
Vision Language Models}

\begin{document}
\sloppy
\twocolumn[
  \icmltitle{BiomedXPro: Prompt Optimization for Explainable Diagnosis with Biomedical
Vision Language Models}



  \icmlsetsymbol{equal}{*}

  \begin{icmlauthorlist}
    \icmlauthor{Kaushitha Silva}{equal,yyy}
    \icmlauthor{Mansitha Eashwara}{equal,yyy}
    \icmlauthor{Sanduni Ubayasiri}{equal,yyy}
    \icmlauthor{Ruwan Tennakoon}{sch}
    \icmlauthor{Damayanthi Herath}{yyy}
  \end{icmlauthorlist}

  \icmlaffiliation{yyy}{Department of Computer Engineering, University of Peradeniya, Peradeniya, Sri Lanka}
  \icmlaffiliation{sch}{School of Computing Technologies, RMIT University, Melbourne, Australia}

  \icmlcorrespondingauthor{Damayanthi Herath}{damayanthiherath@eng.pdn.ac.lk}

  \icmlkeywords{Machine Learning, ICML}

  \vskip 0.3in
]



\printAffiliationsAndNotice{\icmlEqualContribution}

\begin{abstract}
The clinical adoption of biomedical vision-language models is hindered by prompt optimization techniques that produce either uninterpretable latent vectors or single textual prompts. This lack of transparency and failure to capture the multi-faceted nature of clinical diagnosis, which relies on integrating diverse observations, limits their trustworthiness in high-stakes settings. To address this, we introduce \textbf{BiomedXPro} (\underline{Biomed}ical e\underline{X}plainable \underline{Pro}mpts), an evolutionary framework that leverages a large language model as both a biomedical knowledge extractor and an adaptive optimizer to automatically generate a diverse ensemble of interpretable, natural-language prompt pairs for disease diagnosis. Experiments on three biomedical benchmarks show that BiomedXPro consistently outperforms state-of-the-art prompt-tuning methods, particularly in data-scarce few-shot settings. Furthermore, our analysis demonstrates a strong semantic alignment between the discovered prompts and statistically significant clinical features, grounding the model's performance in verifiable concepts. By producing a diverse ensemble of interpretable prompts, BiomedXPro provides a verifiable basis for model predictions, representing a critical step toward the development of more trustworthy and clinically-aligned AI systems.
\end{abstract}
\section{Introduction}
\label{sec:intro}
Accurate and transparent interpretation of biomedical images is fundamental for reliable disease diagnosis. Clinical practioners rely on well-established visual cues such as cellular morphology, tissue architecture, and pathological patterns, combined with domain expertise, to make informed diagnostic decisions. For computer vision systems to achieve clinical acceptance, they must not only demonstrate high predictive accuracy but also provide interpretable outputs that align with established clinical reasoning processes.

Recent advances in Vision-Language Models (VLMs), particularly Contrastive Language-Image Pre-training (CLIP)~\cite{radford_learning_2021}, have demonstrated remarkable potential for bridging visual content with natural language descriptions. Biomedical adaptations such as BiomedCLIP~\cite{zhang_biomedclip_2025} extend these capabilities to medical imaging domains. While CLIP models demonstrate strong zero-shot capabilities, their performance often benefits from prompt optimization tailored to specific tasks. Early attempts relied on manual prompt engineering~\cite{radford_learning_2021}, which was labor-intensive and required domain expertise. To overcome these challenges, gradient-based prompt learning methods such as Context Optimization (CoOp)~\cite{zhou_learning_2022} introduced learnable soft prompts represented as continuous vectors optimized via gradient descent. Biomedical adaptations, including BiomedCoOp~\cite{koleilat_biomedcoop_2025} and XCoOp~\cite{bie_xcoop_2024}, incorporated domain knowledge into CoOp frameworks to improve performance.

Despite these advances, two critical challenges remain for clinical adoption: (1) \textit{prompts used to guide these models are typically optimized as uninterpretable feature vectors, providing no insight into the underlying diagnostic rationale~\cite{koleilat_biomedcoop_2025, bie_xcoop_2024, du_ipo_nodate}}, and (2) \textit{existing methods generally produce singular prompts, restricting their ability to capture the multifaceted nature of clinical observations that practitioners routinely consider.}

Large Language Models (LLMs) present a compelling solution to these limitations. They can generate semantically meaningful, natural-language prompts that encode biomedical knowledge and can be automatically refined to enhance diagnostic performance~\cite{zhou_large_2023, yang_large_nodate}. However, existing LLM-driven prompt optimization techniques still operate largely as black-box systems, with limited mechanisms to ensure clinical transparency or domain relevance.

In this work, we introduce \textit{BiomedXPro}, an evolutionary prompting framework specifically designed for biomedical disease diagnosis. Unlike conventional methods that seek a single optimal prompt, our framework evolves a diverse ensemble of human-readable prompts, each capturing distinct diagnostic observations. These prompts may emphasize specific morphological alterations, tissue organizational patterns, or cellular-level abnormalities that are directly interpretable by clinical practitioners. We leverage LLMs both as biomedical knowledge extractors and as adaptive optimizers that iteratively refine prompts through structured feedback mechanisms, ensuring that the final prompt ensemble captures a comprehensive range of clinically meaningful features.

Our contribution to the field is three-fold. First, we ensure \textbf{interpretability} by anchoring optimized prompts in medically grounded observations, providing direct transparency into the model's decision-making process. Second, by maintaining a \textbf{diverse} ensemble of complementary prompts, \textit{BiomedXPro} mirrors the multi-perspective heuristic employed by clinicians, capturing complex pathological features. Finally, by grounding probabilistic predictions in semantically meaningful concepts, we improve \textbf{clinical trustworthiness}, facilitating the seamless integration of VLM outputs into evidence-based diagnostic workflows. This approach shifts the paradigm from simple accuracy-focused optimization toward a more transparent and clinically-aligned framework for AI-assisted medicine.

\section{Methodology}
\label{sec:methodology}
\subsection{Problem Formulation}

Given a biomedical dataset \(\mathcal{D} = \{(x_i, y_i)\}_{i=1}^N\) comprising medical images \(x_i \in \mathbb{R}^{H \times W \times C}\) and binary labels \(y_i \in \{0,1\}\), we address the challenge of learning interpretable visual concepts for disease classification. The framework utilizes a pre-trained Vision-Language Model (VLM) consisting of an image encoder \(f_\theta: \mathbb{R}^{H \times W \times C} \rightarrow \mathbb{R}^d\) and a text encoder \(g_\phi: \mathcal{T} \rightarrow \mathbb{R}^d\), which project visual and textual inputs into a shared $d$-dimensional embedding space.

The objective is to discover an optimized ensemble of $K$ semantically meaningful prompt pairs:
\begin{equation}
P = \{(p_j^{-}, p_j^{+})\}_{j=1}^K, \quad p_j^{-}, p_j^{+} \in \mathcal{T},
\end{equation}
where $p_j^{+}$ articulates the presence of disease-indicative visual features, and $p_j^{-}$ articulates their absence or a benign state.

For each prompt pair $(p_j^{-}, p_j^{+})$, we define a binary classifier based on the relative cosine similarity in the shared embedding space:
\begin{equation}
h_j(x) = \mathbb{I}\Big[ \text{sim}\big(f_\theta(x), g_\phi(p_j^{+})\big) > \text{sim}\big(f_\theta(x), g_\phi(p_j^{-})\big) \Big],
\end{equation}
where \(\text{sim}(\cdot, \cdot)\) denotes the $\ell_2$-normalized cosine similarity function, and $\mathbb{I}[\cdot]$ is the indicator function.

Ultimately, the global optimization problem is to discover an optimal ensemble $P^*$ (where $|P^*| = K^*$) and a corresponding set of voting weights $W^* = \{w_1, \dots, w_{K^*}\}$ that maximizes an aggregated diagnostic performance metric $\mathcal{M}$ over the dataset $\mathcal{D}$:
\begin{equation}
\begin{aligned}
P^*, W^* &= \underset{P \subset \mathcal{T} \times \mathcal{T}, W}{\arg\max} \; \mathcal{M} \Bigg( \bigg\{ \mathbb{I}\bigg[ \sum_{j=1}^{|P|} w_j h_j(x_i) \\
&\qquad \qquad \qquad > \frac{1}{2}\sum_{j=1}^{|P|} w_j \bigg], y_i \bigg\}_{i=1}^N \Bigg).
\end{aligned}
\end{equation}
Because the search space of natural language $\mathcal{T}$ is discrete, non-differentiable, and practically infinite, discovering $P^*$ via standard gradient-based optimization is intractable. Therefore, the framework approaches this as an evolutionary search problem.

\subsection{BiomedXPro Algorithm}

To navigate this intractable search space, BiomedXPro operationalizes the optimization objective via an evolutionary algorithm. Within this framework, a single prompt pair $(p_j^-, p_j^+)$ serves as the fundamental unit of evolution (the genotype), representing a discrete, testable clinical hypothesis. By iteratively evaluating, archiving, and mutating a population of these prompt pairs, the system systematically converges toward the optimal diagnostic ensemble $P^*$. The complete closed-loop procedure is formalized in Algorithm~\ref{alg:biomedxpro}, with each computational phase detailed in the subsequent subsections. The meta-prompt templates are provided in the Appendix~\ref{app:meta-prompts}.

\begin{algorithm}[htbp]
\caption{BiomedXPro: Evolutionary Prompt Optimization}
\label{alg:biomedxpro}
\begin{algorithmic}[1]
\Require Dataset $\mathcal{D}$, VLM Encoders $f_\theta, g_\phi$, LLM Generator $G_{\text{LLM}}$, Generations $T$, Threshold $\alpha$, Meta-prompt templates $Q_0, Q_t, Q^{\text{crowd}}$.
\Ensure Optimized prompt ensemble $P^*$, Weight vector $W^*$
\State $P^{(0)} \leftarrow \Call{GenerateInitialPopulation}{G_{\text{LLM}}, Q_0}$
\State $\mathcal{A} \leftarrow \emptyset$ \Comment{Initialize evolutionary archive}
\For{$t = 0$ \textbf{to} $T-1$}
    \For{\textbf{each} $(p_j^-, p_j^+) \in P^{(t)}$}
        \State $s_j \leftarrow \Call{EvaluateFitness}{\mathcal{D}, f_\theta, g_\phi, p_j^-, p_j^+}$
        \State $\mathcal{A} \leftarrow \Call{UpdateArchive}{\mathcal{A}, p_j^-, p_j^+, s_j, \alpha}$
    \EndFor
    \State $\mathcal{S}_t \leftarrow \Call{RouletteWheelSample}{\mathcal{A}}$
    \State $\mathcal{S}_t^{\text{sorted}} \leftarrow \Call{SortAscendingByFitness}{\mathcal{S}_t}$
    \State $P^{(t+1)} \leftarrow \Call{Mutation}{G_{\text{LLM}}, Q_t, \mathcal{S}_t^{\text{sorted}}}$
\EndFor
\State $\mathcal{C} \leftarrow \Call{SemanticCluster}{G_{\text{LLM}}, Q^{\text{crowd}}, \mathcal{A}}$
\State $P^* \leftarrow \emptyset, \quad W^* \leftarrow \emptyset$
\For{\textbf{each} $c \in \mathcal{C}$}
    \State $(p^-, p^+, s)^* \leftarrow \Call{SelectElite}{c}$
    \State $P^* \leftarrow P^* \cup \{(p^-, p^+)^*\}$
    \State $W^* \leftarrow W^* \cup \{s^*\}$
\EndFor
\State \Return $P^*, W^*$
\end{algorithmic}
\end{algorithm}

\subsubsection{Initial Population Generation}  
The evolutionary sequence initializes with a meta-prompt $Q_0$ that encodes foundational diagnostic criteria. This meta-prompt instructs a Large Language Model (LLM), denoted as $G_{\text{LLM}}$, to generate the first-generation population of $K_0$ prompt pairs. 

\subsubsection{Fitness Evaluation}  
Each candidate prompt pair $(p_j^{-}, p_j^{+})$ is evaluated against the training set $\mathcal{D}$ to compute a fitness score $s_j$:
\begin{equation}
    s_j = \mathcal{M}\Big( \big\{h_j(x_i), y_i \big\}_{i=1}^N \Big),
\end{equation}
where $\mathcal{M} : \{0,1\}^N \times \{0,1\}^N \to \mathbb{R}$ represents a designated performance metric (e.g., F1-score or balanced accuracy). 

\subsubsection{Evolutionary Archive Update}
Following evaluation, prompt pairs are filtered for inclusion in an evolutionary archive $\mathcal{A}$ via the \texttt{UpdateArchive} operation. This archive serves as an elitist memory buffer that monotonically accumulates high-performing diagnostic concepts across generations:
\begin{equation}
    \mathcal{A} \leftarrow \mathcal{A} \cup \{(p_j^-, p_j^+, s_j) \mid s_j \geq \alpha\}.
\end{equation}
The threshold parameter $\alpha$ acts as a strict minimum viability filter. It is established as either the expected performance of a random classifier (relative to the chosen metric $\mathcal{M}$)

\subsubsection{Probabilistic Sampling and Guided Mutation} 
Unlike stochastic evolutionary operators, BiomedXPro employs the LLM as an intelligent, semantic mutator. At generation $t$, a subset of prompt pairs $\mathcal{S}_t \subset \mathcal{A}$ is sampled via a fitness-proportionate roulette wheel. This probabilistic mechanism balances the exploitation of high-fitness prompts with the exploration of diverse diagnostic concepts.

The sampled subset $\mathcal{S}_t$ is used to dynamically construct the meta-prompt $Q_t$. To counteract the inherent recency bias of LLMs~\cite{yang_large_nodate}, the in-context exemplars within $Q_t$ are first sorted in ascending order of fitness. Furthermore, based on findings from Yang \textit{et al.}~\cite{yang_large_nodate}, fitness scores are normalized to a stable integer range $[60, 90]$. Finally, the generation is conditioned with Chain-of-Thought reasoning~\cite{wei_chain--thought_2023} via prompts such as ``Let's think step-by-step'' to elicit a reasoned next iteration of candidates.

\subsubsection{Semantic Clustering and Elite Selection}
Following $T$ evolutionary generations, a crowding mechanism is applied to the aggregate archive $\mathcal{A}$ to eliminate semantic redundancy. An LLM, guided by a deduplication meta-prompt $Q^{\text{crowd}}$, clusters the candidates into a set of mutually exclusive semantic groups $\mathcal{C} = \{c_1, \dots, c_m\}$ 

Within each cluster $c \in \mathcal{C}$, only the ``elite'' pair possessing the highest fitness score is selected (\texttt{SelectElite}). This yields the final, optimal prompt ensemble $P^*$:
\begin{equation}
P^* \leftarrow \bigcup_{c \in \mathcal{C}} \underset{(p^-, p^+, s) \in c}{\text{arg\,max}} \, s.
\end{equation}

\subsubsection{Final Prediction Formulation}
For inference, the optimized ensemble $P^*$ (where $|P^*| = K^*$) operates as an aggregated committee via weighted majority voting. A given image $x$ is classified as:
\begin{equation}
\hat{y} = \mathbb{I}\left[\sum_{j=1}^{K^*} w_j h_j(x) > \frac{1}{2}\sum_{j=1}^{K^*} w_j\right],
\end{equation}
where the voting weight $w_j = s_j$, ensuring that empirically reliable diagnostic observations exert proportionally greater influence on the final clinical prediction.
\begin{table*}[h]
\centering
\caption{F1-macro percentage results of BiomedXPro compared with baseline methods. Camelyon17-WILDS evaluates domain generalization (train on 3 hospitals, test on unseen). Derm7pt evaluates binary melanoma classification. WBCAtt extends to multiclass via one-vs-rest classifiers.}
\label{tab:results}
\begin{tabular}{l l c c c c c}
\toprule
\textbf{Dataset} & \textbf{Method} & \textbf{1-Shot} & \textbf{2-Shot} & \textbf{4-Shot} & \textbf{8-Shot} & \textbf{16-Shot}\\
\midrule
\multirow{6}{*}{Camelyon17-WILDS}
 & Zero-shot & \multicolumn{5}{c}{41.93} \\
& CoOp        & \textbf{78.79} & 70.76 & 74.15 & 84.25 & 88.47\\
& CoCoOp      & 76.43 & 66.13 & 75.19 & 85.13 & 86.88\\
& BiomedCoOp  & 53.15 & 61.06 & 58.69 & 63.46 & 56.94\\
& XCoOp       & 66.90 & 35.60 & 45.00 & 64.20 & 84.50 \\
\rowcolor{pastelpink} & \textbf{BiomedXPro(Ours)} & 72.06 & \textbf{86.95} & \textbf{90.20} & \textbf{90.87} & \textbf{90.38}\\
\midrule
\multirow{6}{*}{Derm7pt} 
& Zero-shot & \multicolumn{5}{c}{27.86} \\
& CoOp        & 33.91 & 55.88 & 58.70 & 54.92 & 61.38\\
& CoCoOp      & 33.89 & 56.70 & 50.97 & 54.36 & 57.06\\
& BiomedCoOp  & 52.49 & 57.22 & 45.96 & 51.30 & 61.46\\
& XCoOp       & 39.50 & 58.70 & 60.10 & 41.90 & 54.80\\
\rowcolor{pastelpink} &\textbf{BiomedXPro(Ours)} & \textbf{64.54} & \textbf{61.45} & \textbf{60.87} & \textbf{58.51} & \textbf{64.17}\\
\midrule
\multirow{6}{*}{WBCAtt} 
& Zero-shot & \multicolumn{5}{c}{10.50} \\
& CoOp        & 33.08 & 41.74 & 55.10 & 67.81 & \textbf{75.50}\\
& CoCoOp      & 31.86 & 41.91 & 55.96 & 62.02 & 72.19\\
& BiomedCoOp  & 11.24 & 10.83 & 10.48 & 10.57 & 10.91 \\
& XCoOp       & 26.20 & 22.10 & 22.80 & 25.20 & 28.90 \\
\rowcolor{pastelpink}  & \textbf{BiomedXPro(Ours)} & \textbf{41.39} & \textbf{47.31} & \textbf{58.33} & \textbf{69.63} & 72.18 \\
\bottomrule
\end{tabular}
\end{table*}
\section{Experiments and Results}

\subsection{Experimental Setup}
We evaluate on Derm7pt~\cite{kawahara_seven-point_2019}, WBCAtt~\cite{tsutsui_wbcatt_2023}, and Camelyon17-WILDS~\cite{koh_wilds_2021} using BiomedCLIP (ViT-B/16)~\cite{koleilat_biomedcoop_2025} and Gemma 3-27B-it~\cite{team_gemma_2025}. We benchmark few-shot efficiency ($N_c \in \{1,2,4,8,16\}$), out-of-distribution generalization (Camelyon17-WILDS), and clinical semantic alignment against CoOp~\cite{zhou_learning_2022}, CoCoOp~\cite{zhou_conditional_2022}, BiomedCoOp~\cite{koleilat_biomedcoop_2025}, and XCoOp~\cite{bie_xcoop_2024} using the F1-macro score.

The framework evolves an initial population of $K_0=50$ prompt pairs over $T=500$ generations. Fitness is evaluated via inverse binary cross-entropy for 1- and 2-shot regimes, and F1-macro for $\ge 4$-shot regimes.  

\subsection{Results and Analysis}
\paragraph{Quantitative Performance.}  
BiomedXPro consistently outperforms all baselines in the data-scarce few-shot regime (Table~\ref{tab:results}), demonstrating strong generalization across diverse imaging modalities. The relative underperformance of biomedical-specific baselines (BiomedCoOp, XCoOp) indicates that static, single-query LLM prompts artificially constrain the representation space. BiomedXPro circumvents this limitation by continually evolving a dynamic, high-fitness ensemble.

\paragraph{Clinical Relevance.}
To validate semantic alignment, we analyzed the correspondence between the discovered prompt ensembles ($P^*$) and the training data's statistical features (conditional probability $P(\text{class} \mid \text{observation})$). For Derm7pt, the most significant indicator, \textbf{`Linear Irregular' vascular structures} ($P=0.80$), was captured by an elite prompt (F1: 0.65) contrasting a \textit{`regular, linear arrangement'} with a \textit{`chaotic, branching pattern.'} Similarly, for WBCAtt Neutrophil classification, the highly predictive \textbf{unsegmented-band nucleus} ($P=0.85$) was accurately described by an evolved prompt distinguishing a \textit{`predominantly spherical'} nucleus from one that is \textit{`irregular and lobulated.'} This consistent alignment confirms that the evolutionary search natively discovers and articulates statistically significant visual biomarkers. Complete prompt ensembles and conditional probabilities are provided in the supplementary material.

\section{Future Work}
\label{sec:future_work}
Future efforts will focus on extending the framework to native multiclass prompt sets and integrating deterministic semantic metrics for per-iteration diversity maintenance. Furthermore, visual analysis (e.g via Grad-CAM~\cite{selvaraju_grad-cam_2020}) and expert human-in-the-loop validation are essential to ensure model reasoning remains robust and clinically aligned in real-world scenarios.

\section{Conclusion}
\label{sec:conclusion}
We introduce BiomedXPro, an evolutionary framework that generates diverse ensembles of interpretable natural-language prompts for biomedical image classification. Leveraging an LLM as both biomedical knowledge extractor and adaptive optimizer, our method overcomes the limitations of uninterpretable soft prompts and single-prompt systems. BiomedXPro outperforms state-of-the-art prompt-tuning methods in few-shot and domain generalization settings across multiple biomedical datasets. The discovered prompts are not only high-performing but also semantically aligned with statistically significant clinical features, grounding predictions in verifiable concepts. By enhancing interpretability, diversity, and trustworthiness, BiomedXPro advances the safe deployment of vision-language models in clinical practice.

\bibliography{references}
\bibliographystyle{icml2026}

\newpage
\appendix
\onecolumn

\section{Meta-Prompts utilized for each dataset}
\label{app:meta-prompts}
This section presents the exact meta-prompts used for each dataset evaluated in this work.

\subsection{Derm7pt~\cite{kawahara_seven-point_2019} (Melanoma vs Non-melanoma binary classification)}

\begin{itemize}
    \item $Q_0$ --- Initial meta-prompt
    \begin{lstlisting}[caption={Initial meta-prompt for the Derm7pt dataset ($Q_0$).}]

Give 50 distinct textual descriptions of pairs of visual discriminative features to identify whether a dermoscopic image of a skin lesion shows melanoma or not. 
These are the following features an expert would look for: Pigment Network, Blue Whitish Veil, Vascular Structures, Pigmentation, Streaks, Dots and Globules, Regression Structures.
Each description pair must contain two contrasting features: one indicative of benign, and one indicative of a melanoma

Only provide the output as Python code in the following format:
prompts = list[tuple[negative: str, positive: str]]
Let's think step-by-step
    \end{lstlisting}

    \item $Q_t$ --- Mutation meta-prompt
    \begin{lstlisting}[caption={Mutation meta-prompt for the Derm7pt dataset ($Q_t$).}]

The task is to generate distinct textual descriptions pairs of visual discriminative features to identify whether a dermoscopic image of a skin lesion shows melanoma or not.
These are the following features an expert would look for: Pigment Network, Blue Whitish Veil, Vascular Structures, Pigmentation, Streaks, Dots and Globules, Regression Structures.
Each description pair must contain two contrasting features: one indicative of benign, and one indicative of a melanoma

Here are the best performing pairs in ascending order. High scores indicate higher quality visual discriminative features.
Current top 10 prompt pairs:
1. ("....", "...."), Score: 60
2. ..
...

Write 10 new prompt pairs that are different from the old ones and has a score as high as possible, formulate a strategy.
Only provide the output as Python code in the following format: prompts = list[tuple[negative: str, positive: str]]. Let's think step-by-step
    \end{lstlisting}

    \item $Q_{\text{crowd}}$ --- Crowding meta-prompt
    \begin{lstlisting}[caption={Crowding meta-prompt for the Derm7pt dataset ($Q_{\text{crowd}}$).}]

The task is to group textual description pairs of visual discriminative features for melanoma detection in a skin lesion. 
Current Prompt Pairs: Format: <Index. Prompt Pair>
1. ("....", "....")
2. ..
3. ..
...
30. ...

Each pair corresponds to a feature of the same medical concept. Group the prompt pairs that has exactly same observation but differ only in language variations. Give the indexes of the grouped pairs in the output.
Provide the output as follows: list[list[index:int]]. Make sure to include all pairs in the output, even if they are not grouped with others.
Let's think step by step.
    \end{lstlisting}
\end{itemize}

\subsection{WBCAtt~\cite{tsutsui_wbcatt_2023} (Peripheral Blood Smears: Multiclass Cell-type classification)}

The following meta-prompts were utilized for the WBCAtt dataset, which focuses on multiclass cell-type classification in peripheral blood smears. For each one-vs-rest binary classifier, the variables \texttt{\{BINARY\_LABEL\}} and \texttt{\{other\_labels\_str\}} are replaced with the specific cell-type being evaluated. The prompt leverages a set of defining attributes for each white blood cell type, which are stored in the \texttt{MEDICAL\_CONCEPTS\_MAPPING} dictionary. These attributes are:

\begin{itemize}
\item \textbf{Basophil}: Nucleus=Segmented, NC Ratio=Low, Granularity=Yes, Color=Blue/Black (dense)
\item \textbf{Eosinophil}: Nucleus=Segmented, NC Ratio=Low, Granularity=Yes, Color=Red
\item \textbf{Lymphocyte}: Nucleus=Unsegmented, NC Ratio=High, Granularity=No, Size=Small
\item \textbf{Monocyte}: Nucleus=Unsegmented, NC Ratio=Low, Granularity=No
\item \textbf{Neutrophil}: Nucleus=Segmented, NC Ratio=Low, Granularity=Yes, Color=Blue
\end{itemize}

\vspace*{2em}
\begin{itemize}
    \item $Q_0$ --- Initial meta-prompt
     \begin{lstlisting}[caption={Initial meta-prompt for the WBCAtt dataset ($Q_0$).}]

Give 50 distinct textual descriptions of pairs of visual discriminative features to identify whether the peripheral blood cell is a {BINARY_LABEL} or not. Other cell types include {other_labels_str}. 
These are the following features an expert would look for: Cell Size, Cell Shape, Nucleus Shape, Nuclear-Cytoplasmic Ratio, Chromatin-Density, Cytoplasm-Vacuole, Cytoplasm-Texture, Cytoplasm-Color, Granule-Type, Granule-Color, Granularity
{MEDICAL_CONCEPTS_MAPPING[BINARY_LABEL]}

Only provide the output as Python code in the following format: prompts = list[tuple[negative: str, positive: str]]. Let's think step-by-step
    \end{lstlisting}

    \item $Q_t$ --- Mutation meta-prompt
    \begin{lstlisting}[caption={Mutation meta-prompt for the WBCAtt dataset ($Q_t$).}]

The task is to generate distinct textual descriptions of pairs of visual discriminative features to identify whether the peripheral blood cell is a {BINARY_LABEL} or not. Other cell types include {other_labels_str}. 
These are the following features an expert would look for: Cell Size, Cell Shape, Nucleus Shape, Nuclear-Cytoplasmic Ratio, Chromatin-Density, Cytoplasm-Vacuole, Cytoplasm-Texture, Cytoplasm-Color, Granule-Type, Granule-Color, Granularity
{MEDICAL_CONCEPTS_MAPPING[BINARY_LABEL]}

Here are the best performing pairs in ascending order. High scores indicate higher quality visual discriminative features.

Current top 10 prompt pairs:
1. ("....", "...."), Score: 60
2. ..
...

Write 10 new prompt pairs that are different to from the old ones and has a score as high as possible. Formulate a strategy",
Only provide the output as Python code in the following format: prompts = list[tuple[negative: str, positive: str]]. Let's think step-by-step
    \end{lstlisting}

    \item $Q_{\text{crowd}}$ --- Crowding meta-prompt
    \begin{lstlisting}[caption={Crowding meta-prompt for the WBCAtt dataset ($Q_{\text{crowd}}$).}]

The task is to group textual description pairs of visual discriminative features for white blood cell classification. 
Current Prompt Pairs: Format: <Index. Prompt Pair>
1. ("....", "....")
2. ..
3. ..
...
30. ...

Each pair corresponds to a feature of the same medical concept. Group the prompt pairs that has exactly same observation but differ only in language variations. Give the indexes of the grouped pairs in the output.
Provide the output as follows: list[list[index:int]]. Make sure to include all pairs in the output, even if they are not grouped with others.
Let's think step by step.
    \end{lstlisting}
\end{itemize}

\subsection{Camelyon17-WILDS~\cite{koh_wilds_2021}(Histopathology: Binary Classification)}

\begin{itemize}
    \item $Q_0$ --- Initial meta-prompt
    \begin{lstlisting}[caption={Initial meta-prompt for the Camelyon17-WILDS dataset ($Q_0$).}]

Give 50 distinct textual descriptions of pairs of visual discriminative features to identify whether the central region of a histopathological image patch contains tumor tissue or not. The patch is extracted from an H&E-stained whole-slide image of a lymph node section. Only provide the output as Python code in the following format: prompts = list[tuple[negative: str, positive: str]]
    \end{lstlisting}

    \item $Q_t$ --- Mutation meta-prompt
    \begin{lstlisting}[caption={Mutation meta-prompt for the Camelyon17-WILDS dataset ($Q_t$).}]

The task is to generate distinct textual descriptions pairs of visual discriminative features to identify whether the central region of a histopathological image patch contains tumor tissue or not. The patch is extracted from an H&E-stained whole-slide image of a lymph node section.
Here are the best performing pairs in ascending order. High scores indicate higher quality visual discriminative features.

Current top 10 prompt pairs:
1. ("....", "...."), Score: 60
2. ..
...

Write 10 new prompt pairs that are different to from the old ones and has a score as high as possible.
Only provide the output as Python code in the following format: prompts = list[tuple[negative: str, positive: str]]. Let's think step-by-step
    \end{lstlisting}

    \item $Q_{\text{crowd}}$ --- Crowding meta-prompt
    \begin{lstlisting}[caption={Crowding meta-prompt for the Camelyon17-WILDS dataset ($Q_{\text{crowd}}$).}]
The task is to group textual description pairs of visual discriminative features for tumor detection in histopathology. 
Current Prompt Pairs: Format: <Index. Prompt Pair>
1. ("....", "....")
2. ..
3. ..
...
30. ...
Each pair corresponds to a feature of the same medical concept. Group the prompt pairs that has exactly same observation but differ only in language variations. Give the indexes of the grouped pairs in the output.
Provide the output as follows: list[list[index:int]]. Make sure to include all pairs in the output, even if they are not grouped with others. Let's think step by step
    \end{lstlisting}
\end{itemize}

\section{Sample LLM CoT Strategy}

This section presents sample Chain-of-Thought (CoT) responses generated by the Gemma 3-27b-it LLM, demonstrating its reasoning process for our meta-prompts. \cref{lst:strategy} and \cref{lst:strategy_2} show two distinct examples of the strategy formulated by the LLM in response to the optimization meta-prompt ($Q_t$) for the Camelyon17-WILDS dataset. An example of the reasoning and final output for the crowding meta-prompt ($Q^{\text{crowd}}$) is provided in \cref{lst:crowding_reasoning}.

\begin{lstlisting}[caption={Example of LLM's reasoning for the strategy as a response for the optimization meta prompt},label={lst:strategy}]
Strategy to Enhance Prompt Fitness:
1. Highlight histopathological features that unambiguously separate benign/reactive from tumor tissue.
2. Use precise, quantifiable terms based on standard diagnostic criteria.
3. Introduce new descriptors-nuclear size & contour, chromatin texture, necrosis, apoptosis, cell cohesion, architectural disruption, cytoplasmic granularity, and hallmark patterns (e.g., starry sky, rosettes).
4. Craft negative prompts around the absence or benign manifestation of these features, and positive prompts around their abnormal or tumor-associated presentations.
5. Avoid repeating any feature pairs from the existing top-10 list.
\end{lstlisting}

\begin{lstlisting}[caption={Example of LLM's reasoning for the strategy as a response for the optimization meta prompt},label={lst:strategy_2}]
Okay, let's break down a strategy to generate high-scoring prompt pairs for identifying tumor tissue in histopathological images.
1. Analyze Existing High-Scoring Pairs:
Specificity: The best pairs aren't just "normal" vs. "abnormal." They pinpoint specific features. "Abnormal cytoplasmic staining" is better than just "abnormal cells."
Visual Discriminability: The features must be readily observable in an H&E stain. Things like molecular markers are out of scope.
Tumor-Related Features: The features directly relate to hallmarks of cancer - abnormal growth, nuclear changes, necrosis, invasion, etc.
Contrast: The negative and positive prompts are clearly contrasting. "No mitotic figures" vs. "Frequent mitotic figures" is strong.
Lymph Node Context: Since the images are from lymph nodes, features related to lymph node architecture (lymphatic invasion, reactive follicles) are relevant.
2. Brainstorming Feature Categories:
Based on the above, let's consider categories of features to focus on:
Nuclear Features: Size, shape, chromatin, nucleoli, membrane irregularities. These are very important in cancer diagnosis.
Cytoplasmic Features: Staining intensity, granularity, presence of inclusions.
Growth Patterns: How cells are arranged (sheets, clusters, single cells, palisading).
Microenvironment: Inflammation, fibrosis, vascular changes, necrosis.
Invasion: Evidence of tumor cells breaking through boundaries (capsule, vessels, lymphatics).
Differentiation: Features suggesting the cell type or degree of differentiation (e.g., plasmacytoid differentiation).
3. Generating Prompt Pairs:
I'll aim for pairs that are:
Specific and visually discriminable.
Contrasting.
Relevant to lymph node pathology.
Different from the existing top 10.
\end{lstlisting}

\begin{lstlisting}[
    frame=single,
    caption={Example of the LLM’s reasoning and output in response to the crowding meta-prompt.},
    label={lst:crowding_reasoning},
    basicstyle=\small\ttfamily,
    breaklines=true
]
Let's think step by step. I will analyze the prompt pairs and group them based on identical observations with language variations.

* Group 1: No plasma cells identified/observed/present
    * 3: ('No plasma cells are identified.' , 'Numerous plasma cells with eccentric nuclei and abundant cytoplasm.')
    * 12: ('No plasma cells are identified.' , 'Numerous plasma cells with eccentric nuclei are present.')
    * 14: ('No plasma cells are observed.' , 'Increased number of plasma cells, some with Dutcher bodies.')
    * 17: ('No plasma cells are identified.' , 'Increased number of plasma cells, some with Dutcher bodies.')
    * 18: ('No plasma cells are present.' , 'Increased number of plasma cells with eccentric nuclei.')
    * 22: ('No plasma cells are observed.' , 'Abundant plasma cells with eccentric nuclei.')
    * 24: ('No plasma cells are identified.' , 'Abundant plasma cells with eccentric nuclei are present.')

* Group 2: Few reactive lymphocytes vs. Numerous reactive lymphocytes
    * 4: ('Few scattered reactive lymphocytes.' , 'Sheets of atypical lymphocytes with brisk proliferation.')
    * 6: ('Few reactive lymphocytes are present.' , 'Dense infiltrate of atypical lymphocytes and plasma cells.')
    * 8: ('Few reactive lymphocytes are present.' , 'Numerous reactive lymphocytes and plasma cells are observed.')
    * 13: ('Few reactive lymphocytes present.' , 'Numerous reactive lymphocytes with plasmacytoid differentiation.')
    * 25: ('Few reactive lymphocytes are present.' , 'Numerous reactive lymphocytes are present, forming rosettes around tumor cells.')
    * 27: ('Few reactive lymphocytes.' , 'Abundant reactive lymphocytes and atypical lymphoid cells.')
    * 30: ('Few reactive lymphocytes present.' , 'Numerous reactive lymphocytes forming rosettes around tumor cells.')

* Group 3: Lymphocyte maturity
    * 9: ('Mature lymphocytes.' , 'Immunoblasts and plasmablasts.')
    * 20: ('Lymphocytes are small and mature-appearing.' , 'Lymphocytes are large and immature-appearing with prominent nucleoli.')
    * 29: ('Lymphocytes are small and mature.' , 'Lymphocytes are large and immature with prominent nucleoli.')

* Group 4: Stroma cellularity
    * 11: ('Stroma is edematous with minimal cellularity.' , 'Stroma is densely cellular with a myxoid appearance.')
    * 28: ('Stroma is hypocellular with minimal collagen.' , 'Stroma is hypercellular with abundant collagen deposition.')

* Group 5: Atypical plasma cells
    * 19: ('No atypical plasma cells are identified.' , 'Presence of atypical plasma cells with prominent nucleoli.')

* Group 6: Inflammatory infiltrate
    * 15: ('Inflammatory infiltrate is minimal and lymphocytic.' , 'Dense inflammatory infiltrate with neutrophils and eosinophils.')
    * 21: ('Minimal inflammatory infiltrate.' , 'Dense inflammatory infiltrate with numerous plasma cells.')

* Group 7: Eosinophils
    * 16: ('No eosinophils are present.' , 'Increased number of eosinophils infiltrating the tumor.')

* Group 8: Emperipolesis
    * 10: ('No emperipolesis is observed.' , 'Frequent emperipolesis is seen within tumor cells.')

* Group 9: Germinal Centers
    * 26: ('Few reactive germinal centers.' , 'Numerous large and atypical germinal centers with prominent nucleoli.')

* Group 10: Infiltration of atypical cells
    * 1: ('No infiltration of atypical cells.' , 'Infiltration of atypical cells disrupting normal lymph node architecture.')
    * 5: ('No infiltration of atypical cells into surrounding tissues.' , 'Infiltration of atypical cells into surrounding tissues, disrupting normal structures.')

* Group 11: Cell Arrangement
    * 2: ('Cells arranged in a normal lymphatic pattern.' , 'Cells arranged in sheets or solid nests.')

* Group 12: Background Lymphoid Aggregates
    * 23: ('Background shows normal lymphoid aggregates.' , 'Background shows diffuse sheets of atypical cells.')

* Group 13: Immunocytes
    * 7: ('Immunocytes are rare.' , 'Abundant immunocytes are present, forming rosettes.')

--- Final Output ---
[
    [3, 12, 14, 17, 18, 22, 24],
    [4, 6, 8, 13, 25, 27, 30],
    [9, 20, 29],
    [11, 28],
    [19],
    [15, 21],
    [16],
    [10],
    [26],
    [1, 5],
    [2],
    [23],
    [7]
]
\end{lstlisting}
\section{Prompts Learnt}
This section presents the final prompt ensembles ($P^*$) generated by BiomedXPro. We provide the prompts learned in two distinct experimental settings: the full-shot scenario, where the F1-macro score was used as the fitness metric, and the 1-shot scenario, where fitness was evaluated using the inverted Binary Cross-Entropy (BCE).

\subsection{Derm7pt}
The prompts learnt for the 1-shot and full-shot scenarios are listed in \cref{tab:derm7pt_prompts} and \cref{tab:derm7pt_1shot_prompts} respectively. 
\begin{table}[htbp]
  \centering
  \caption{Prompts Learned for the Derm7pt full dataset: F1-macro utilized as the fitness function}
  \label{tab:derm7pt_prompts}
  \begin{tabular}{|p{0.4\textwidth}|p{0.4\textwidth}|c|}
    \hline
    \textbf{Absence or Normal Appearance ($p_j^-$)} & \textbf{Disease-Indicative Observations ($p_j^+$)} & \textbf{Fitness Score ($s_j$)} \\
    \hline
    Pigmentation is limited to brown tones with sharp borders. & Pigmentation is variegated with shades of brown, black, blue, and grey, with ill-defined borders. & 0.6741 \\
    \hline
    Pigmentation is homogenous and limited to the epidermis. & Pigmentation is heterogeneous and extends into the dermis, causing a bluish hue. & 0.6724 \\
    \hline
    Blue-whitish veil is absent. & Blue-whitish veil is present and associated with ulceration. & 0.6667 \\
    \hline
    Pigmentation is limited to the periphery with a normal skin color center. & Pigmentation extends to the center of the lesion with shades of black, grey, and blue. & 0.6565 \\
    \hline
    Dots and globules have a smooth, rounded appearance. & Dots and globules are angular, irregular, and exhibit pseudopods. & 0.6543 \\
    \hline
    Pigmentation is light brown or tan. & Pigmentation is dark brown or black. & 0.6535 \\
    \hline
    Vascular structures show a regular, linear arrangement following skin creases. & Vascular structures display a chaotic, branching pattern with irregular vessel diameters. & 0.6523 \\
    \hline
    Vascular structures are primarily hairpin-shaped telangiectasias. & Vascular structures show branching vessels with irregular caliber and occasional hemorrhages. & 0.6506 \\
    \hline
    Pigmentation is predominantly brown with subtle variations. & Pigmentation includes shades of black, grey, and blue, with abrupt color changes. & 0.6419 \\
    \hline
    Vascular structures are primarily small, regularly spaced capillaries. & Vascular structures demonstrate large, dilated vessels and hemorrhages. & 0.6392 \\
    \hline
    The lesion exhibits a smooth, symmetrical surface with no evidence of ulceration or scarring. & The lesion exhibits an irregular, asymmetrical surface with ulceration, crusting, and surrounding inflammation. & 0.635 \\
    \hline
    Regression structures are absent or minimal, with slight lightening of the skin. & Regression structures are significant, with a scarred, fibrotic appearance and possible ulceration. & 0.6324 \\
    \hline
    The lesion exhibits a smooth, even surface texture. & The lesion demonstrates surface elevation, ulceration, or a nodular appearance. & 0.6322 \\
    \hline
    Streaks are fine and follow the natural skin creases. & Streaks are thick, irregular, and abruptly change direction, suggesting rapid growth. & 0.6304 \\
    \hline
  \end{tabular}
\end{table}

\begin{table}[htbp]
  \centering
  \caption{Prompts Learned for the 1-Shot Derm7pt Case: Inverted BCE is utilized for the fitness function}
  \label{tab:derm7pt_1shot_prompts}
  \begin{tabular}{|p{0.4\textwidth}|p{0.4\textwidth}|c|}
    \hline
    \textbf{Absence or Normal Appearance ($p_j^-$)} & \textbf{Disease-Indicative Observations ($p_j^+$)} & \textbf{Fitness Score ($s_j$)} \\
    \hline
    Blue whitish veil is absent or minimal. & Blue whitish veil is pronounced and covers a significant portion of the lesion. & 1.0 \\
    \hline
    Blue whitish veil is absent or very subtle. & Blue whitish veil is prominent and widespread, obscuring the underlying structures. & 1.0 \\
    \hline
    Dots are evenly distributed and small. & Dots are clustered and large, with irregular shapes. & 1.0 \\
    \hline
    Dots and globules are small, round, and uniformly pigmented. & Dots and globules are large, irregular, and exhibit color variation. & 1.0 \\
    \hline
    Blue-whitish veil is absent or minimal, with no associated changes. & Blue-whitish veil is dense and associated with surrounding inflammation. & 1.0 \\
    \hline
  \end{tabular}
\end{table}

\subsection{WBCAtt}
The prompts learned for the \textbf{Neutrophil} class in the full-shot and 1-shot scenarios are shown in \cref{tab:wbcatt_neutrophil_prompts} and \cref{tab:wbcatt_neutrophil_1shot_prompts}, respectively.

\begin{table}[htbp]
  \centering
  \caption{Prompts Learned for Neutrophil Classification (Full-Shot Scenario): F1-macro as Fitness Function}
  \label{tab:wbcatt_neutrophil_prompts}
  \begin{tabular}{|p{0.4\textwidth}|p{0.4\textwidth}|c|}
    \hline
    \textbf{Negative Class Observation ($p_j^-$)} & \textbf{Neutrophil-Indicative Observation ($p_j^+$)} & \textbf{Fitness Score ($s_j$)} \\
    \hline
    The cell shape is consistently round with a smooth perimeter. & The cell shape is often irregular and slightly distorted due to the segmented nucleus. & 0.7945 \\
    \hline
    Cytoplasm is clear and homogenous, lacking distinct features. & Cytoplasm is abundant and contains fine, pale lilac granules. & 0.7710 \\
    \hline
    The nuclear shape is predominantly spherical with a smooth outline. & The nuclear shape is irregular and lobulated, with a clearly indented outline. & 0.7565 \\
    \hline
    Granules, if present, are sparse and difficult to discern. & Granules are abundant, easily visible, and fill a significant portion of the cytoplasm. & 0.7487 \\
    \hline
    The cell shape is generally round or slightly irregular. & The cell exhibits a more polymorphic shape, often displaying pseudopods or extensions. & 0.7467 \\
    \hline
    The cell shape is consistently oval or spherical, maintaining a regular form. & The cell shape is highly variable and often irregular, exhibiting pseudopod extensions. & 0.7418 \\
    \hline
    The cell shape is generally spherical or slightly irregular. & The cell exhibits an elongated or flattened shape. & 0.7358 \\
    \hline
    Granules, if present, are small and indistinct. & Granules are large, prominent, and easily identifiable. & 0.7340 \\
    \hline
    The cell size is similar to that of a small lymphocyte. & The cell size is significantly larger, comparable to a monocyte. & 0.7242 \\
    \hline
    The cell displays a uniform, homogenous cytoplasm without any noticeable texture. & The cell displays a granular cytoplasm with a slightly textured appearance. & 0.7213 \\
    \hline
    Granules, if present, are sparse and appear as small, indistinct dots. & Granules are numerous, prominent, and exhibit a distinct, segmented appearance. & 0.7147 \\
    \hline
    The cell exhibits a uniform, smooth cytoplasmic border. & The cell exhibits an irregular cytoplasmic border with occasional projections. & 0.7133 \\
    \hline
    The cell exhibits a high nuclear-cytoplasmic ratio, with the nucleus occupying most of the cell volume. & The cell displays a balanced nuclear-cytoplasmic ratio, with a substantial amount of cytoplasm. & 0.7109 \\
    \hline
    Granules, if present, are very few and evenly distributed. & Granules are numerous and clustered in specific areas of the cytoplasm. & 0.7062 \\
    \hline
  \end{tabular}
\end{table}

\begin{table}[htbp]
  \centering
  \caption{Prompts Learned for Neutrophil Classification (1-Shot Scenario): Inverted BCE as Fitness Function}
  \label{tab:wbcatt_neutrophil_1shot_prompts}
  \begin{tabular}{|p{0.4\textwidth}|p{0.4\textwidth}|c|}
    \hline
    \textbf{Negative Class Observation ($p_j^-$)} & \textbf{Neutrophil-Indicative Observation ($p_j^+$)} & \textbf{Fitness Score ($s_j$)} \\
    \hline
    Basophils granules contain histamine and heparin, giving them a metachromatic staining property. & Neutrophils granules contain enzymes like myeloperoxidase and lysozyme, staining specifically with neutral dyes. & 1.0000 \\
    \hline
    Lymphocytes have a condensed, hyperchromatic nucleus. & Neutrophils have a segmented nucleus with varying degrees of chromatin condensation in each lobe. & 1.0000 \\
    \hline
    Basophils often have a U- or S-shaped nucleus due to the abundance of granules. & Neutrophils maintain a consistently segmented nuclear shape, even with granule presence. & 1.0000 \\
    \hline
    Eosinophils have granules that are rich in major basic protein and eosinophil cationic protein. & Neutrophils have granules containing elastase and collagenase. & 1.0000 \\
    \hline
    Basophils have granules that stain intensely with basic dyes. & Neutrophils have granules that stain weakly with both acidic and basic dyes. & 1.0000 \\
    \hline
    Basophils exhibit a high nuclear-cytoplasmic ratio, with a relatively small amount of cytoplasm. & Neutrophils have a low nuclear-cytoplasmic ratio, with a significant amount of cytoplasm. & 0.9942 \\
    \hline
    Lymphocytes have a consistently round nucleus. & Neutrophils have a segmented nucleus, typically with 3-5 lobes. & 0.9515 \\
    \hline
    Monocytes often present with a folded or irregular nuclear shape. & Neutrophils consistently display a segmented nucleus with distinct lobes. & 0.5331 \\
    \hline
    Basophils have intensely staining, dark purple granules that often fill the cytoplasm. & Neutrophils have granules that are less intensely stained and more evenly distributed within the cytoplasm. & 0.2721 \\
    \hline
    Eosinophils typically have a bilobed nucleus, although variations exist. & Neutrophils are characterized by a multi-lobed nucleus, usually with 3-5 lobes. & 0.2497 \\
    \hline
    Cells with a large, round nucleus and minimal cytoplasm are often lymphocytes. & Neutrophils are identified by their segmented nucleus and relatively large amount of cytoplasm. & 0.2492 \\
    \hline
    Eosinophilic cytoplasm is indicative of eosinophils. & The cytoplasm is pale blue or slightly pink. & 0.2332 \\
    \hline
    Monocytes typically exhibit a bean-shaped or kidney-shaped nucleus. & Neutrophils are defined by a multi-lobed, segmented nucleus. & 0.2013 \\
    \hline
  \end{tabular}
\end{table}

\subsection{Camelyon17-WILDS}
The prompts learned in both the 1-shot and 16-shot scenarios are shown in \cref{tab:camelyon17_1shot_prompts} and \cref{tab:camelyon17_16shot_prompts}, respectively. 

\begin{table}[htbp]
  \centering
  \caption{Prompts Learned for Camelyon17-WILDS (1-Shot Scenario): Inverted BCE as Fitness Function}
  \label{tab:camelyon17_1shot_prompts}
  \begin{tabular}{|p{0.4\textwidth}|p{0.4\textwidth}|c|}
    \hline
    \textbf{Absence or Normal Appearance ($p_j^-$)} & \textbf{Disease-Indicative Observations ($p_j^+$)} & \textbf{Fitness Score ($s_j$)} \\
    \hline
    Cells arranged in a normal lymphatic pattern. & Cells arranged in sheets or solid nests. & 0.9998 \\
    \hline
    No infiltration of atypical cells. & Infiltration of atypical cells disrupting normal lymph node architecture. & 0.9998 \\
    \hline
    No plasma cells are identified. & Numerous plasma cells with eccentric nuclei and abundant cytoplasm. & 0.9994 \\
    \hline
    Few scattered reactive lymphocytes. & Sheets of atypical lymphocytes with brisk proliferation. & 0.9992 \\
    \hline
    Immunocytes are rare. & Abundant immunocytes are present, forming rosettes. & 0.9988 \\
    \hline
    Mature lymphocytes. & Immunoblasts and plasmablasts. & 0.9987 \\
    \hline
    No emperipolesis is observed. & Frequent emperipolesis is seen within tumor cells. & 0.9986 \\
    \hline
    Stroma is edematous with minimal cellularity. & Stroma is densely cellular with a myxoid appearance. & 0.9984 \\
    \hline
    No eosinophils are present. & Increased number of eosinophils infiltrating the tumor. & 0.9980 \\
    \hline
    Inflammatory infiltrate is minimal and lymphocytic. & Dense inflammatory infiltrate with neutrophils and eosinophils. & 0.9980 \\
    \hline
    No atypical plasma cells are identified. & Presence of atypical plasma cells with prominent nucleoli. & 0.9979 \\
    \hline
    Background shows normal lymphoid aggregates. & Background shows diffuse sheets of atypical cells. & 0.9973 \\
    \hline
    Few reactive germinal centers. & Numerous large and atypical germinal centers with prominent nucleoli. & 0.9971 \\
    \hline
    Lymphocytes show low proliferative activity. & Lymphocytes exhibit brisk proliferation with many dividing cells. & 0.9966 \\
    \hline
    Normal vasculature with thin walls. & Thickened and irregular vasculature with increased blood flow. & 0.9962 \\
    \hline
    Low nuclear-to-cytoplasmic ratio in lymphocytes. & High nuclear-to-cytoplasmic ratio in tumor cells. & 0.9953 \\
    \hline
    Germinal centers are small and inactive. & Germinal centers are large and hyperactive with numerous mitotic figures. & 0.9947 \\
    \hline
    Minimal inflammatory infiltrate. & Dense inflammatory infiltrate with lymphocytes and plasma cells. & 0.9946 \\
    \hline
    Adipose tissue is present in normal distribution. & Adipose tissue is compressed or replaced by tumor. & 0.9946 \\
    \hline
    Clear cytoplasm in lymphocytes. & Darkly stained, irregular cytoplasm in tumor cells. & 0.9941 \\
    \hline
    Lymphocytes are small and uniform. & Tumor cells are large and pleomorphic. & 0.9941 \\
    \hline
    No mitoses observed. & Frequent mitoses with atypical mitotic figures. & 0.9940 \\
    \hline
    Stroma is collagenous and hypocellular. & Stroma is fibrotic and hypercellular with increased mitotic activity. & 0.9938 \\
    \hline
    Few plasma cells are seen. & Numerous plasma cells with atypical morphology. & 0.9937 \\
    \hline
  \end{tabular}
\end{table}

\begin{table}[htbp]
  \centering
  \caption{Prompts Learned for Camelyon17-WILDS (16-Shot Scenario): F1-macro as Fitness Function}
  \label{tab:camelyon17_16shot_prompts}
  \begin{tabular}{|p{0.4\textwidth}|p{0.4\textwidth}|c|}
    \hline
    \textbf{Absence or Normal Appearance ($p_j^-$)} & \textbf{Disease-Indicative Observations ($p_j^+$)} & \textbf{Fitness Score ($s_j$)} \\
    \hline
    No increase in mitotic activity. & Increased mitotic activity. & 0.9193 \\
    \hline
    No increase in eosinophilic cytoplasm. & Increased eosinophilic cytoplasm in tumor cells. & 0.8982 \\
    \hline
    Normal follicular architecture is preserved. & Disrupted follicular architecture with loss of polarity. & 0.8904 \\
    \hline
    Lymphocytes are small and round with dense nuclei. & Lymphocytes are large and atypical with prominent nucleoli. & 0.8859 \\
    \hline
    Benign inflammatory infiltrate. & Aggressive, atypical inflammatory infiltrate. & 0.8837 \\
    \hline
    No prominent nucleolus. & Large and prominent nucleoli within tumor cells. & 0.8822 \\
    \hline
    No increase in nuclear size. & Marked increase in nuclear size (nuclear pleomorphism). & 0.8806 \\
    \hline
    No atypical cells present. & Presence of atypical cells with pleomorphic nuclei. & 0.8763 \\
    \hline
    No fibrosis. & Increased stromal fibrosis surrounding tumor nests. & 0.8748 \\
    \hline
    No increase in interfollicular proliferation. & Increased interfollicular proliferation of atypical cells. & 0.8663 \\
    \hline
    No high-grade nuclear features. & Presence of high-grade nuclear features. & 0.8627 \\
    \hline
    Cells exhibit normal size and shape. & Cells exhibit significant variation in size and shape (anisocytosis and pleomorphism). & 0.8532 \\
    \hline
    No increased number of macrophages. & Increased number of macrophages infiltrating the tissue. & 0.8490 \\
    \hline
    Stroma with minimal collagen deposition. & Desmoplastic stroma with abundant collagen deposition. & 0.8476 \\
    \hline
    No spindle cell morphology. & Predominantly spindle cell morphology. & 0.8298 \\
    \hline
    No perivascular lymphocytic cuffing. & Perivascular lymphocytic cuffing around tumor nests. & 0.8297 \\
    \hline
    No increased nuclear-to-cytoplasmic ratio. & Increased nuclear-to-cytoplasmic ratio in tumor cells. & 0.8278 \\
    \hline
    No syncytial formation. & Syncytial formation of tumor cells. & 0.8233 \\
    \hline
    No evidence of extracellular mucin. & Presence of extracellular mucin surrounding tumor cells. & 0.8225 \\
    \hline
  \end{tabular}
\end{table}
\clearpage
\section*{Clinical Relevance Analysis}

To provide a deeper, evidence-based validation of the prompts' clinical relevance, we ground our analysis in the statistical features of the training data. To create a quantitative measure of importance for the visual features in a dataset, we compute the conditional probability $P(\text{class} \mid \text{observation})$. This is calculated by dividing the number of images of a given class that exhibit a specific observation by the total number of images where that observation is present, across the training set. This probability serves as a data-driven proxy for the diagnostic significance of a visual feature.

\begin{table}[h]
    \centering
    \caption{Conditional probabilities of melanoma given specific clinical observations, calculated on the Derm7pt training dataset (N=412).}
    \label{tab:supp_clinical_relevance_derm7pt}
    \begin{tabular}{p{5.5cm} c}
    \hline
    \textbf{Clinical Observation} & \textbf{P(Melanoma $\vert$ Obs.)} \\
    \hline
    Vascular Structures: Linear Irregular & 0.80 \\
    Blue-Whitish Veil: Present           & 0.59 \\
    Regression Structures: Combinations  & 0.58 \\
    Pigment Network: Atypical            & 0.55 \\
    Streaks: Irregular                   & 0.53 \\
    Pigmentation: Diffuse Irregular      & 0.48 \\
    Vascular Structures: Dotted          & 0.48 \\
    Dots and Globules: Irregular         & 0.47 \\
    Regression Structures: White Areas   & 0.42 \\
    Pigmentation: Localized Irregular    & 0.33 \\
    Regression Structures: Blue Areas    & 0.33 \\
    \hline
    \end{tabular}
\end{table}

\subsection*{Derm7pt Analysis}
For the \textbf{Derm7pt} dataset, we calculated these probabilities for key dermatoscopic features associated with melanoma. The results are presented in \cref{tab:supp_clinical_relevance_derm7pt}. A detailed comparison between these statistical findings and the learned full-shot prompts (presented in \cref{tab:derm7pt_prompts}) reveals a strong and consistent alignment:

\begin{itemize}
    \item \textbf{Vascular Structures:} The most predictive feature, \textit{‘Linear Irregular’ vascular structures} ($P=0.80$), was captured by the high-fitness prompt pair (F1: 0.6523) that contrasts a \textit{'regular, linear arrangement following skin creases'} with a \textit{'chaotic, branching pattern with irregular vessel diameters.'}

    \item \textbf{Blue-Whitish Veil:} The second most significant feature, the presence of a \textit{‘Blue-Whitish Veil’} ($P=0.59$), was identified by a high-scoring prompt (F1: 0.6667) distinguishing its absence from when it is \textit{'present and associated with ulceration.'}

    \item \textbf{Regression Structures \& Streaks:} Other key indicators like \textit{regression structures} ($P=0.58$) and \textit{irregular streaks} ($P=0.53$) were also identified, with prompts learning to contrast \textit{'absent or minimal'} structures with \textit{'significant, with a scarred, fibrotic appearance,'} and \textit{'fine, natural'} streaks with those that are \textit{'thick, irregular, and abruptly change direction.'}

    \item \textbf{Pigment Network:} Notably, while \textit{‘Atypical Pigment Network’} was a strong statistical indicator ($P=0.55$), prompts describing nuanced aspects of pigmentation consistently achieved the highest F1-scores (up to 0.6741), such as the top-performing pair contrasting \textit{'limited to brown tones with sharp borders'} with being \textit{'variegated with shades of brown, black, blue, and grey.'}
\end{itemize}

\subsection*{WBCAtt Analysis}
A similar analysis was performed for the WBCAtt dataset, focusing on the one-vs-rest classification of Neutrophils. We calculated the conditional probabilities for key cytological observations, presented in \cref{tab:supp_clinical_relevance_wbcatt}. The following is a detailed comparison between these statistics and the prompts learned in the full-shot scenario (presented in \cref{tab:wbcatt_neutrophil_prompts}).

\begin{itemize}
    \item \textbf{Granules:} The statistical data identifies the presence of \textit{‘Small’} and \textit{‘Pink’} granules as perfect predictors ($P=1.00$). This is directly captured by the second-highest scoring prompt (F1: 0.7710), which contrasts a \textit{'clear and homogenous'} cytoplasm with one that is \textit{'abundant and contains fine, pale lilac granules.'} Multiple other high-fitness prompts also focus on the abundance and visibility of granules.

    \item \textbf{Nucleus Shape:} The \textit{‘Unsegmented-band’} nucleus shape is a highly significant statistical feature ($P=0.85$). The framework learned this concept, generating a high-fitness prompt (F1: 0.7565) that distinguishes a \textit{'predominantly spherical'} nucleus from one that is \textit{'irregular and lobulated, with a clearly indented outline.'}

    \item \textbf{Complex Feature Learning:} Interestingly, while \textit{‘Irregular’} cell shape is a weak statistical predictor on its own ($P=0.32$), the highest-scoring learned prompt (F1: 0.7945) is about cell shape. It learns a more complex, conditional feature by contrasting a \textit{'consistently round'} shape with one that is \textit{'often irregular and slightly distorted due to the segmented nucleus.'} This shows the framework's ability to discover nuanced, multi-part features that are more discriminative than the simple statistical correlations.
\end{itemize}

This alignment demonstrates that BiomedXPro effectively identifies and articulates the key cytological features necessary for cell type classification.
\begin{table}[h]
    \centering
    \caption{Conditional probabilities of a cell being a Neutrophil given specific cytological observations, calculated on the WBCAtt training dataset.}
    \label{tab:supp_clinical_relevance_wbcatt}
    \begin{tabular}{p{5.5cm} c}
    \hline
    \textbf{Cytological Observation} & \textbf{P(Neutrophil $\vert$ Obs.)} \\
    \hline
    Granule Type: Small                  & 1.00 \\
    Granule Colour: Pink                 & 1.00 \\
    Nucleus Shape: Unsegmented-band      & 0.85 \\
    Granularity: Yes                     & 0.43 \\
    Cytoplasm Colour: Light blue         & 0.43 \\
    Cytoplasm Texture: Clear             & 0.40 \\
    Nuclear-Cytoplasmic Ratio: Low       & 0.37 \\
    Cell Size: Big                       & 0.36 \\
    Chromatin Density: Densely           & 0.35 \\
    Nucleus Shape: Segmented-multilobed  & 0.34 \\
    Cytoplasm Vacuole: No                & 0.34 \\
    Cell Shape: Round                    & 0.32 \\
    Cell Shape: Irregular                & 0.32 \\
    Cell Size: Small                     & 0.28 \\
    Nucleus Shape: Segmented-bilobed     & 0.20 \\
    Cytoplasm Vacuole: Yes               & 0.15 \\
    \hline
    \end{tabular}
\end{table}

\subsection*{Limitation on Multiclass Classification}
As mentioned in the main paper, while the one-vs-rest scheme is effective for extending our binary framework to multiclass problems, it can lead to the generation of over-specialized negative prompts ($p_j^-$). Instead of learning a general description for a "non-target" class, the evolutionary process often discovers that the most effective strategy is to describe a single, high-contrast competing class. The prompts learned for the \textbf{Eosinophil} class in the WBCAtt full-shot scenario provide a clear example of this behavior.

\begin{table}[h]
  \centering
  \caption{Prompts Learned for Eosinophil Classification (Full-Shot Scenario)}
  \label{tab:wbcatt_eosinophil_prompts}
  \begin{tabular}{|p{0.4\textwidth}|p{0.4\textwidth}|c|}
    \hline
    \textbf{Negative Class Observation ($p_j^-$)} & \textbf{Eosinophil-Indicative Observation ($p_j^+$)} & \textbf{Fitness Score ($s_j$)} \\
    \hline
    Neutrophils exhibit granules that stain neutrally, appearing a faint lilac color. & Eosinophil granules are intensely stained with eosin, resulting in a bright red-orange color. & 0.8311 \\
    \hline
    Monocytes exhibit a dull, greyish cytoplasm. & Eosinophils display a cytoplasm with a pale pink or slightly orange hue. & 0.8247 \\
    \hline
    Basophils have granules that stain metachromatically, appearing purple-black. & Eosinophil granules stain specifically with eosin, resulting in a bright orange-red color. & 0.8179 \\
    \hline
    Basophils display dark blue or purple granules that often overwhelm the cytoplasm. & Eosinophil granules are characteristically bright pink or red-orange, contrasting with the basophilic granules of basophils. & 0.8092 \\
    \hline
    Monocytes display a folded or indented nuclear membrane. & Eosinophil nuclei are generally round or oval with a smooth, regular membrane. & 0.8079 \\
    \hline
    Monocytes exhibit a "cloud-like" or "ground glass" appearance to their cytoplasm. & Eosinophil cytoplasm is filled with distinct, refractile granules, lacking the diffuse appearance of monocytes. & 0.8015 \\
    \hline
    The cytoplasm of a lymphocyte is usually scant and clear. & Eosinophil cytoplasm is plentiful and filled with characteristic red-orange granules. & 0.7971 \\
    \hline
  \end{tabular}
\end{table}

A review of the highest-scoring prompts (presented in \cref{tab:wbcatt_eosinophil_prompts}) shows that the negative prompts are not general descriptions of blood cells, but are instead specific definitions of other cell types:
\begin{itemize}
    \item The top-performing prompt (F1: 0.8311) uses a description of a \textbf{Neutrophil} for its negative case: \textit{'Neutrophils exhibit granules that stain neutrally, appearing a faint lilac color.'}
    \item The second-highest scoring prompt (F1: 0.8247) uses a \textbf{Monocyte} as the negative exemplar: \textit{'Monocytes exhibit a dull, greyish cytoplasm.'}
    \item The third-highest scoring prompt (F1: 0.8179) uses a \textbf{Basophil}: \textit{'Basophils have granules that stain metachromatically, appearing purple-black.'}
\end{itemize}

This analysis shows that the framework effectively converts the one-vs-rest problem into a series of highly discriminative one-vs-one classifiers. While this is a powerful strategy for achieving high performance, it creates classifiers that may be brittle and less generalizable to the full spectrum of other classes. This finding further motivates the future work discussed in the main paper: extending the framework to natively handle a multiclass prompt set.
\section{Ablation studies}
\subsection{Effect of prompt selection criteria. }

\begin{figure}[h]
    \centering
    \includegraphics[width=0.5\linewidth]{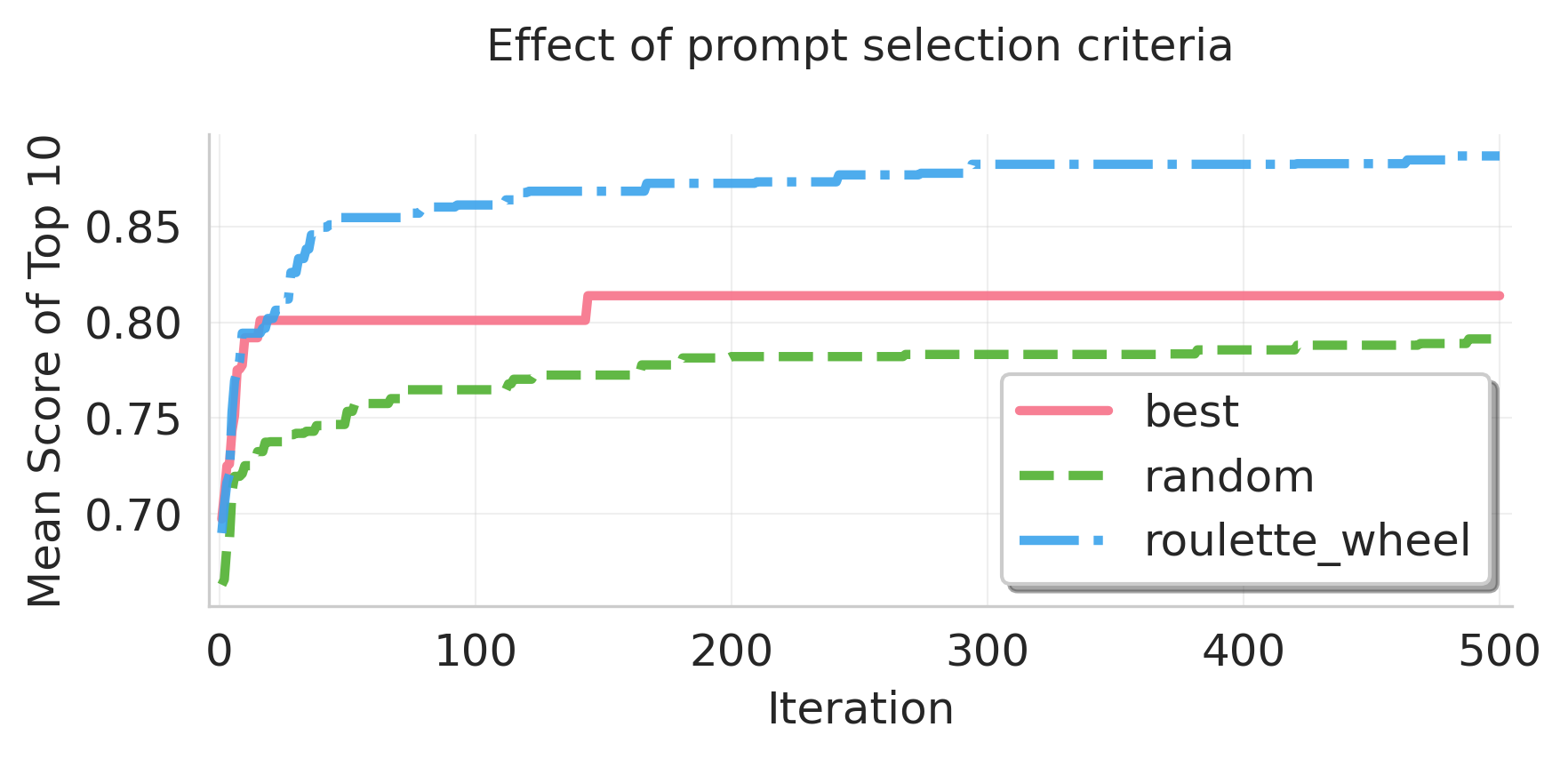}
    \caption{Effect of prompt selection strategy on convergence. Our proposed \textit{Roulette Wheel} selection achieves the best balance of exploration and exploitation, leading to more stable and consistent performance gains compared to \textit{Best-N} and \textit{Random} selection.}
    \label{fig:selection_curves}
\end{figure}

We analyze the impact of the strategy for sampling mutation candidates by comparing our proposed \textit{Roulette Wheel Selection} against two baselines: deterministic \textit{Best-N} and uniform \textit{Random} selection. As shown in \cref{fig:selection_curves}, these baselines represent extremes of the exploration-exploitation trade-off. \textit{Best-N} selection focuses purely on exploitation, leading to premature convergence in local optima, while \textit{Random} selection's pure exploration provides an inefficient and erratic search signal to the LLM. Our approach effectively balances this trade-off by probabilistically favoring higher-fitness candidates while maintaining sufficient randomness for exploration, resulting in the stable and consistent performance gains shown in the learning curves.

We analyze the impact of the strategy used to sample prompt pairs from the archive $\mathcal{A}$ for the next generation's mutation. We compare our proposed \textit{Roulette Wheel Selection} against two baselines: (1) \textit{Best-N Selection}, which deterministically chooses the top-$K$ prompts by fitness, and (2) \textit{Random Selection}, which samples uniformly.

As illustrated in Figure~\ref{fig:selection_curves}, these strategies represent extremes in the exploration-exploitation trade-off. \textit{Best-N Selection} focuses purely on exploitation, which frequently leads to premature convergence. By repeatedly feeding the same high-scoring prompts to the LLM, the optimization process often stalls in local optima until stochasticity in the LLM's output provides an escape. Conversely, \textit{Random Selection} prioritizes exploration but often feeds the LLM low-fitness prompts, diluting the quality of information available for refinement and resulting in an erratic and inefficient search process.

Our proposed \textit{Roulette Wheel Selection} effectively balances this trade-off. By probabilistically favoring higher-fitness candidates, it provides the LLM with a strong, informative basis for generating superior prompts, while the inherent randomness ensures sufficient exploration to avoid getting stuck. The learning curves in Figure~\ref{fig:selection_curves} empirically validate this, demonstrating that our approach achieves more stable and consistent performance improvement throughout the evolutionary process.

\subsection{Impact of generation size per iteration.}

\begin{figure}[h]
    \centering
    \includegraphics[width=0.5\linewidth]{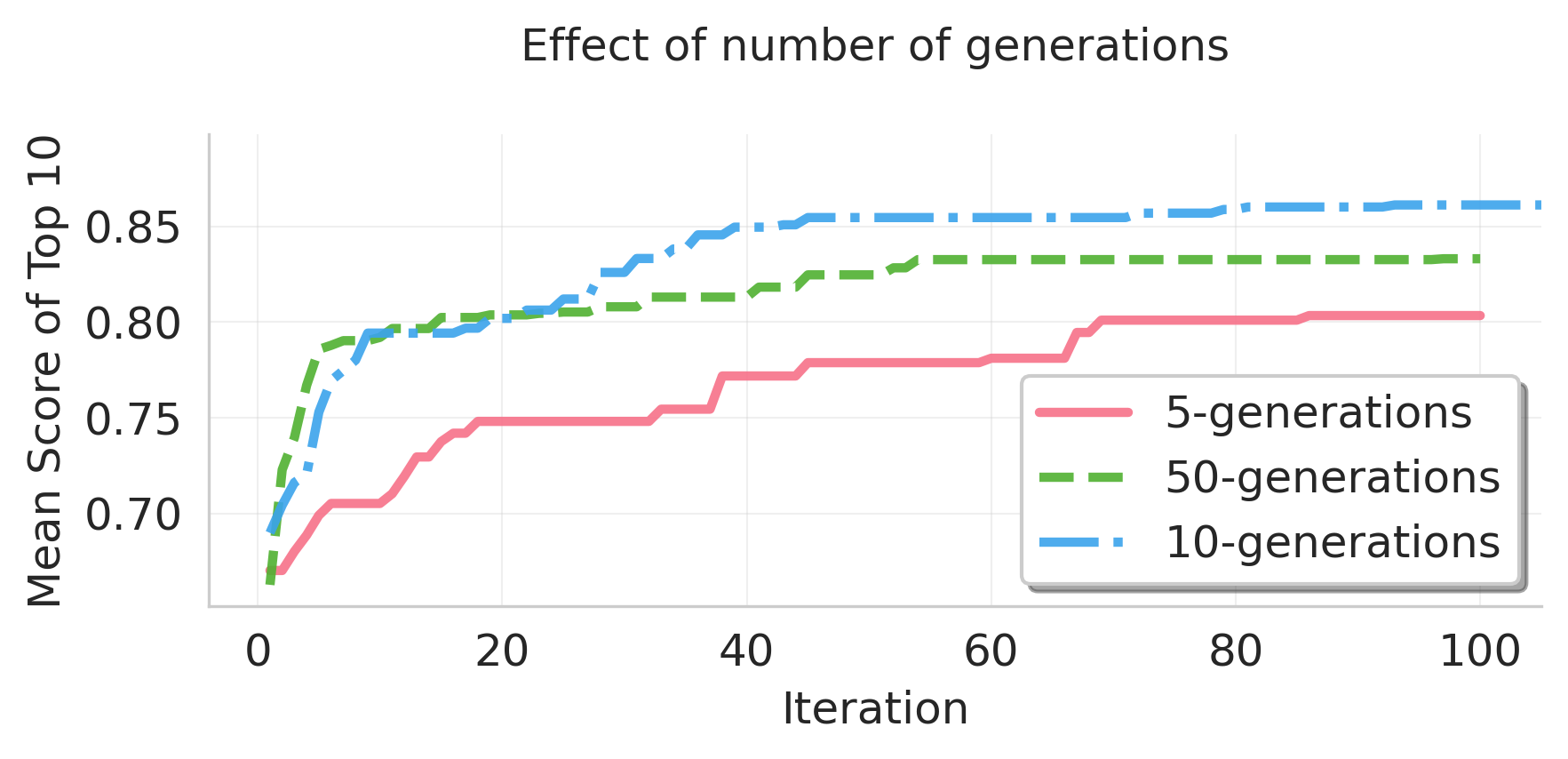}
    \caption{Impact of generation size per iteration on performance. The 10-pair setting (our choice) provides the optimal balance, outperforming 5 pairs and 50 pairs.}
    \label{fig:generation_size_ablation}
\end{figure}

We also analyze the number of prompt pairs generated per mutation step, comparing our choice of 10 against smaller (5) and larger (50) sets. As shown in \cref{fig:generation_size_ablation}, generating only 5 pairs leads to slow convergence, while a large set of 50 causes performance to plateau quickly after a strong start. While the precise reason for this degradation with a large generation size is unclear, this finding is consistent with prior work by Yang et al.~\cite{yang_large_nodate}, who also observed that a moderate number of generated instructions was optimal. Based on these findings, we utilize a generation size of 10, as it provides a favorable balance between refinement diversity and optimization stability while being more cost-effective in terms of token usage.

\subsection{Effect of chain-of-thought prompting.}

\begin{figure}[h]
    \centering
    \includegraphics[width=0.5\linewidth]{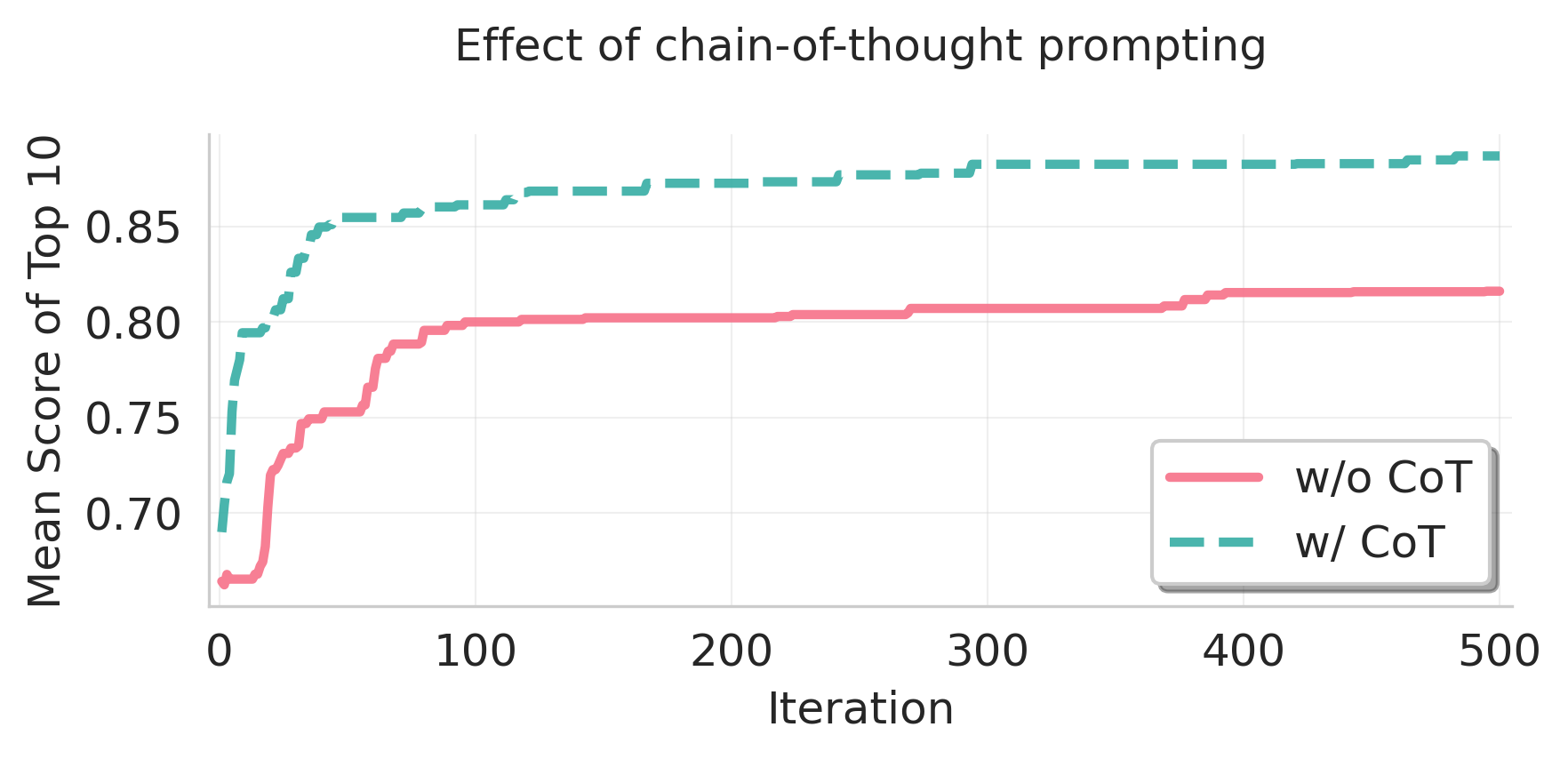}
    \caption{Effect of Chain-of-Thought (CoT) prompting, which consistently outperforms a direct-instruction baseline.}
    \label{fig:cot_ablation}
\end{figure}

We investigated the effect of Chain-of-Thought (CoT) prompting by adding an instruction for the LLM to first "formulate a strategy" and "Let's think step-by-step" for improvement before generating new candidates. As shown in \cref{fig:cot_ablation}, this addition leads to consistently better performance. Examples of CoT generated by the LLM are provided in the supplementary material.

\subsection{Effect of initial population size.}
\begin{figure}[h]
\centering
\includegraphics[width=0.5\linewidth]{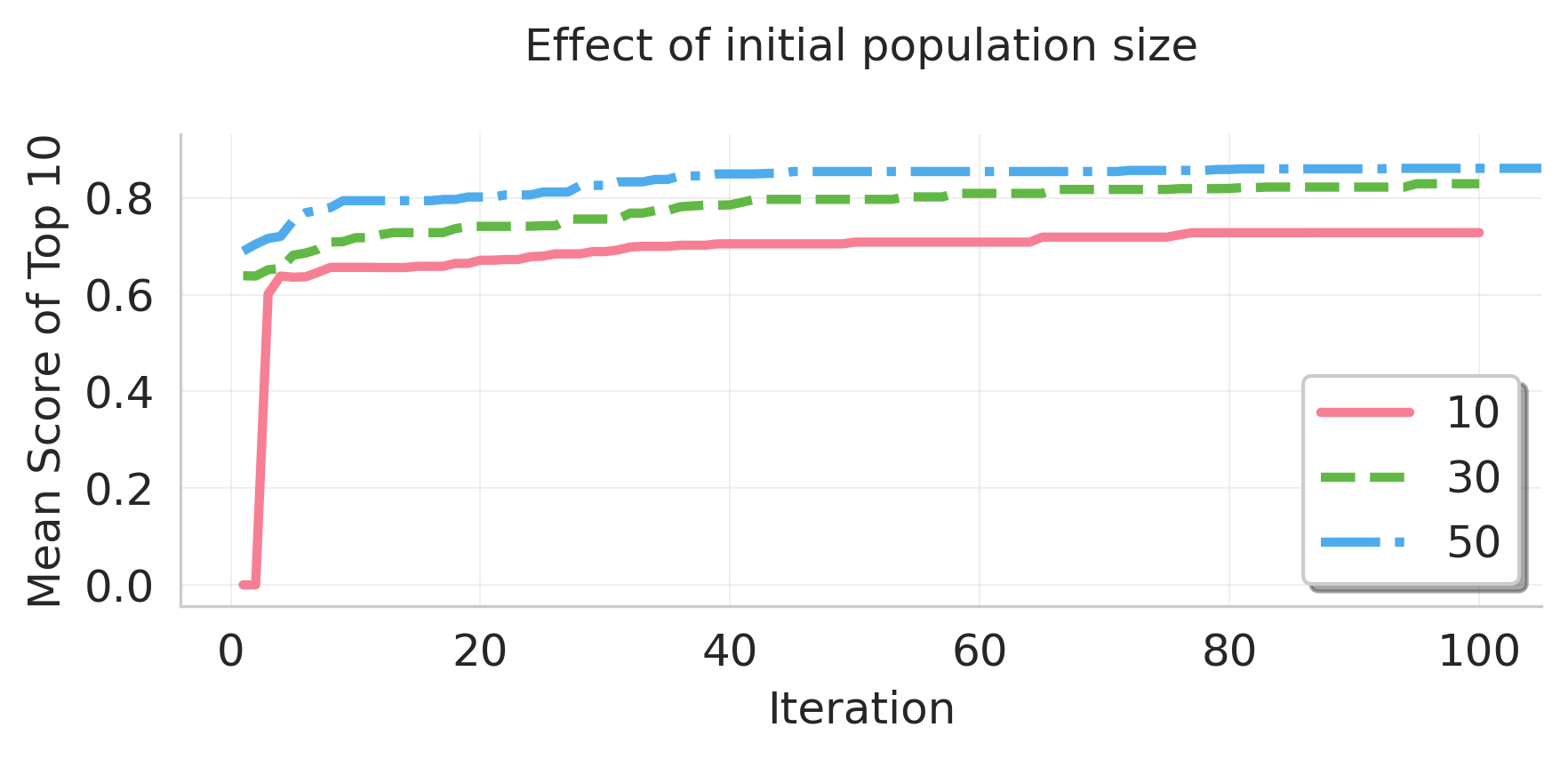}
\caption{Effect of initial population size on performance. A larger and more diverse initial population (50) consistently outperforms smaller sets}
\label{fig:initial_population_size}
\end{figure}

We compare initial sizes $K_0$ of 10, 30, and 50, and find that a larger initial population of 50 consistently outperforms the smaller sets (\cref{fig:initial_population_size}). This suggests that a broader initial semantic space provides a more robust foundation for the optimization, preventing premature convergence on a narrow range of concepts. We did not explore sizes beyond 50 due to the practical limitations of our LLM, which struggled to reliably generate a large and exact number of distinct prompts in a single query.

\end{document}